# Channel-independent and sensor-independent stimulus representations

David N. Levin
*Department of Radiology, University of Chicago, Chicago, IL 60637*
*(d-levin@uchicago.edu)*

This paper shows how a machine, which observes stimuli through an uncharacterized, uncalibrated channel and sensor, can glean machine-independent information (i.e., channel- and sensor-independent information) about the stimuli. This is possible if the following two conditions are satisfied by the observed stimulus and by the observing device, respectively: 1) the stimulus' trajectory in the space of all possible configurations has a well-defined local velocity covariance matrix; 2) the observing device's sensor state is invertibly related to the stimulus state. The first condition guarantees that statistical properties of the stimulus time series endow the stimulus configuration space with a differential geometric structure (a metric and parallel transfer procedure), which can then be used to represent relative stimulus configurations in a coordinate-system-independent manner. This requirement is satisfied by a large variety of physical systems, and, in general, it is expected to be satisfied by stimuli with velocity distributions varying smoothly across stimulus state space. The second condition means that the machine defines a specific coordinate system on the stimulus state space, with the nature of that coordinate system depending on the machine's channels and detectors. Thus, machines with different channels and sensors "see" the same stimulus trajectory through state space, but in different machine-specific coordinate systems. It is shown that this requirement is almost certainly satisfied by any device that measures more than $2d$ independent properties of the stimulus, where $d$ is the number of stimulus degrees of freedom. Taken together, the two conditions guarantee that the observing device can record the stimulus time series in its machine-specific coordinate system and then derive coordinate-system-independent (and, therefore, machine-independent) representations of relative stimulus configurations. The resulting description is an "inner" property of the stimulus time series in the sense that it does not depend on extrinsic factors like the observer's choice of a coordinate system in which the stimulus is viewed (i.e., the observer's choice of channels and sensors). In other words, the resulting description is an intrinsic property of the evolution of the "real" stimulus that is "out there" broadcasting energy to the observer. This methodology is illustrated with analytic examples and with a numerically simulated experiment. In an intelligent sensory device, this kind of representation "engine" could function as a "front end" that passes channel/sensor-independent stimulus representations to a pattern recognition module. After a pattern recognizer has been trained in one of these devices, it could be used without change in other devices having different channels and sensors.

## I. INTRODUCTION

Conventional sensory devices typically detect energy from an evolving physical stimulus and then use the resulting signal time series to reconstruct the temporal evolution of the stimulus state. If that state is numerically represented by the stimulus' configuration in a specific coordinate system, it can only be recovered if the sensory device's response has been calibrated with respect to that coordinate system. For example, if a camera is observing a moving particle, the particle's position in the laboratory coordinate system can be only be recovered if the camera has a known response to the particle at known positions in that coordinate system. This calibration is typically done by exposing the system to a "test pattern" of known stimulus states and by recording the relationship between those states and the device's sensor states. Such calibration "tables" make it possible for two machines to compensate for differences in their channels and sensors and thereby create identical representations of identical stimulus states.

Remarkably, different humans seem to create similar representations of the world without using such explicit calibration procedures. Specifically, two individuals with similar life experiences tend to represent the world in the same way despite apparently uncompensated differences in the channels and sensors through which they observe the world. For instance, two observers tend to produce similar representations of auditory stimuli despite unknown differences in their external and middle ears, their cochleae, and the neural architectures of their primary auditory cortices. The results of this biological "experiment" suggest the possibility of building machines that observe stimuli through different channels and sensors and that independently create identical stimulus representations without using any physical knowledge of those channels or sensors.

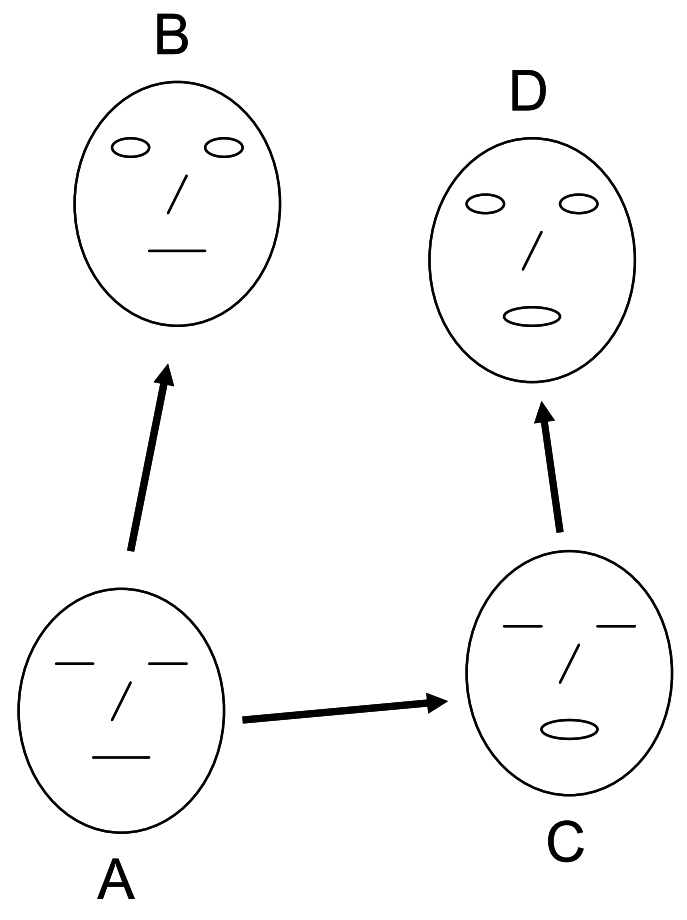

FIG. 1. Consider a two dimensional space populated by faces characterized by the extent to which the mouth and eyes are open. Given two faces (*A* and *B*) that are related by a small transformation ( $A \rightarrow B$ , called an "eye-opening" transformation) and given another face (*C*) that differs from *A* by another small transformation ( $A \rightarrow C$ , called a "mouth-opening" transformation), observers will often judge a particular transformation of *C*, namely $C \rightarrow D$ , to be perceptually equivalent to $A \rightarrow B$ . In other words, they will perceive the stimuli as being related by the analogy: *A:B* = *C:D*.

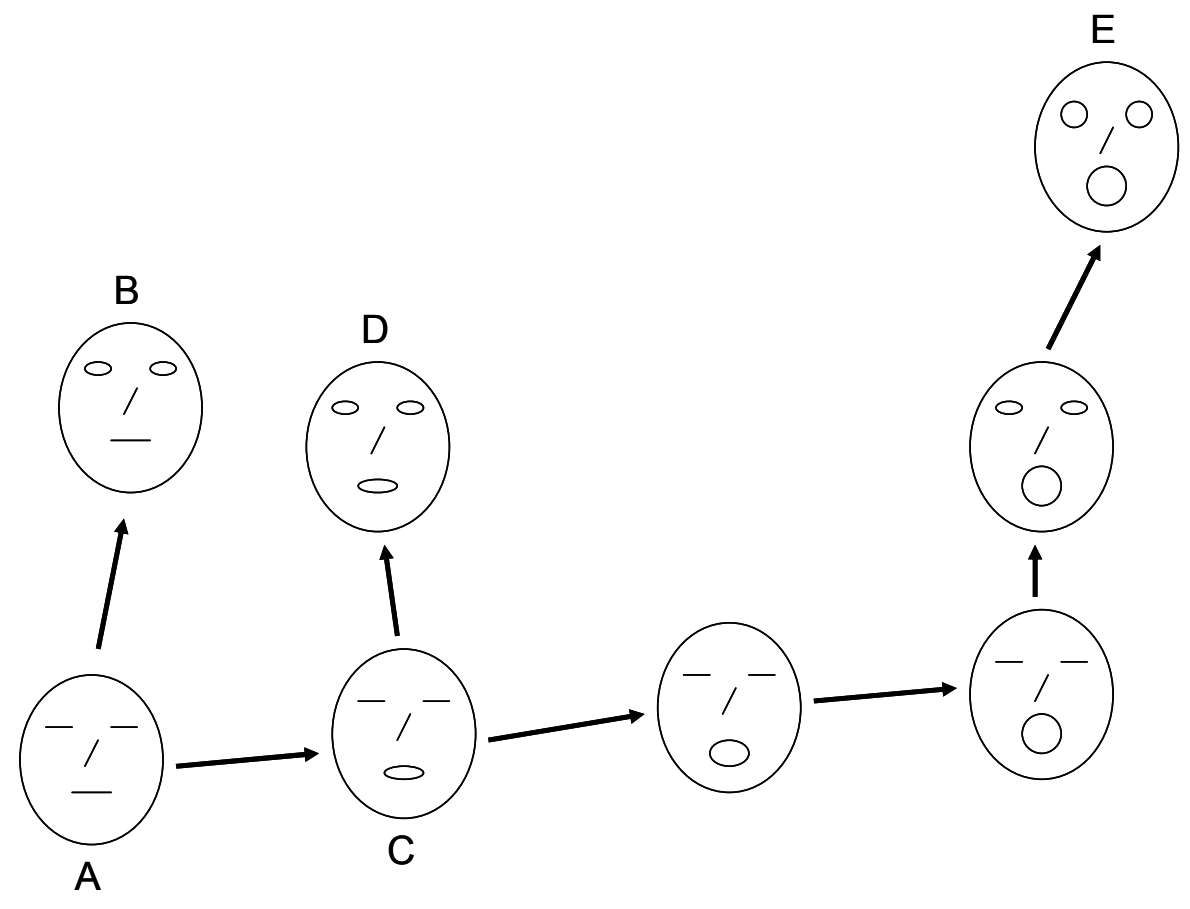

FIG. 2. "Distant" stimuli can be related by describing a path connecting them as a sequence of perceptually defined stimulus transformations. For example, an observer may describe *E* as the stimulus that is produced by starting at *A* and performing three transformations perceptually equivalent to the mouth-opening transformation $A \rightarrow C$ , followed by two transformations perceptually equivalent to the eye-opening transformation $A \rightarrow B$ .

Human perceptual characteristics suggest some general properties of the stimulus representations that such a machine might use. Without the use of an explicit calibration procedure, it is not possible to derive an *absolute* stimulus representation (i.e., a representation that describes the stimulus configuration in a specific physical coordinate system), and, in fact, humans infrequently describe the world in such absolute terms. Instead, people usually describe stimulus perceptions in a *relative* manner; i.e., in relation to their perceptions of other stimuli. Thus, if an individual is asked to describe a light or a shape or a sound, he/she is likely to compare it to his/her perception of other lights or shapes or sounds. People may make certain "primitive" perceptual judgments that are used to construct the perceived relationships between stimuli. For example, in most circumstances, observers have a strong sense about whether two stimuli are the same or not; i.e., whether stimulus *A* = *B* or not. Likewise, observers often feel that stimulus "analogies" are true or false. For instance (Fig. 1), suppose that stimuli *A*, *B*, *C*, and *D* differ by small stimulus transformations. Observers will often have a definite sense about whether the relationship between *A* and *B* is the same (or is not the same) as the relationship between *C* and *D*; i.e., whether *A:B* = *C:D* or not. A series of such perceived analogies can be concatenated in order to describe relationships between two "distant" stimuli, by specifying how one stimulus can be transformed into the other[1]. Figure 2 shows an example in which an observer describes stimulus *E* as being related to stimuli *A*, *B*, and C by a succession of analogous stimulus transformations: "*E* is the stimulus that is produced by starting with stimulus *A* and performing three transformations perceptually equivalent to the $A \rightarrow C$ transformation, followed by two transformations perceptually equivalent to the $A \rightarrow B$ transformation". In principle, an observer, who perceives local stimulus analogies at each point in stimulus space, can use this technique to describe the paths connecting any two stimuli of interest. The set of all such descriptions of relative stimulus locations provides a kind of "map of the world" that can be used to navigate through stimulus space. In fact, it comprises a relative representation of the stimulus space that is analogous to a geographic representation of a city, which specifies how to navigate between any two points in it without specifying the latitudes and longitudes of those points.

If the brain uses stimulus analogies to impose order on the "world" of stimuli, it seems likely that these analogies are derived from past sensory history, rather than being "hard-wired" from birth. For example, there is evidence that young children "learn to see" and that adults with no previous visual experience usually do not sense meaningful relationships between visual stimuli[2]. The latter phenomenon is illustrated by a number of neurological case histories in which congenital cataracts were excised from the eyes of a blind adult with the intention of allowing the patient to see for the first time[3]. In each case, when the post-operative bandages were removed, the patient invariably experienced an inchoate mixture of meaningless visual patterns, presumably because he/she never learned how to impose order on visual information. *In short, a consideration of human perception suggests that an intelligent sensory machine should use its "life history" of previously-encountered signals in order to "learn" primitive stimulus analogies, which can then be used to represent stimuli by their relative locations in stimulus space.*

One can devise many ways to use earlier signals to define a set of stimulus analogies. However, the approximate universality of human perception suggests that people derive such analogies in a particular way: namely, in a way that is *invariant* under channel- and sensor-induced transformations of their primary sensory cortical states. To illustrate, suppose that stimuli *A*, *B*, *C*, and *D* create complex spatial distributions of neuronal electrical discharges (denoted $x_A$, $x_B$, $x_C$, and $x_D$) in the primary sensory cortex of observer $Ob$. The same stimuli will produce different three-dimensional electrical current distributions ($x'_A$, $x'_B$, $x'_C$, and $x'_D$) in the sensory cortex of a second observer $Ob'$, who has a different sensory apparatus. Yet, despite these differences, if the two observers have had similar life histories and if $Ob$ senses $x_A : x_B = x_C : x_D$, then $Ob'$ tends to independently conclude $x'_A : x'_B = x'_C : x'_D$. In other words, the stimulus analogy tends to be invariant under the transformation $x \rightarrow x'$ that maps the sensor states of $Ob$ onto those of $Ob'$. A related fact is that each individual's perceptions of stimuli tend to be independent of the channel through which he/she views those stimuli. This fact was strikingly illustrated by experiments in which subjects wore goggles that created severe geometric distortions and inversions of the observed scenes[4]. Although each subject initially perceived the distortion of the new channel, his/her perceptions of the world tended to return to the pre-experimental baseline after several weeks of constant exposure to familiar stimuli seen through the goggles. These observations suggest the following generalization. Consider an individual who has a certain history of stimulus exposure and "learns" to perceive relationships (e.g., $x_A : x_B = x_C : x_D$) among certain stimulus-induced sensory cortex states. If that person is exposed to a similar history of stimuli through a new observational channel, he/she will relearn relationships among sensory cortex states with the result that the altered sensory cortex states $x'$ induced by stimuli through the new channel will be perceived to be related to one another in the same way (e.g., $x'_A : x'_B = x'_C : x'_D$) as the corresponding unaltered cortex states $x$ were related before the channel changed. *In summary: the above-described phenomena suggest that the brain uses its past experience to derive stimulus analogies that are invariant in the presence of channel-induced and sensor-induced transformations of recently-encountered sensory cortex states.* In this paper, we describe a machine that represents stimuli by their relative locations in state space, using stimulus analogies that are derived from the machine's history of sensor states[5]. Following the above discussion, we require the stimulus analogies to be derived so that they are invariant under systematic transformations of those sensor states. As shown in Section II, this requirement of invariance is such a strong constraint that it largely determines what the stimulus analogies must be.

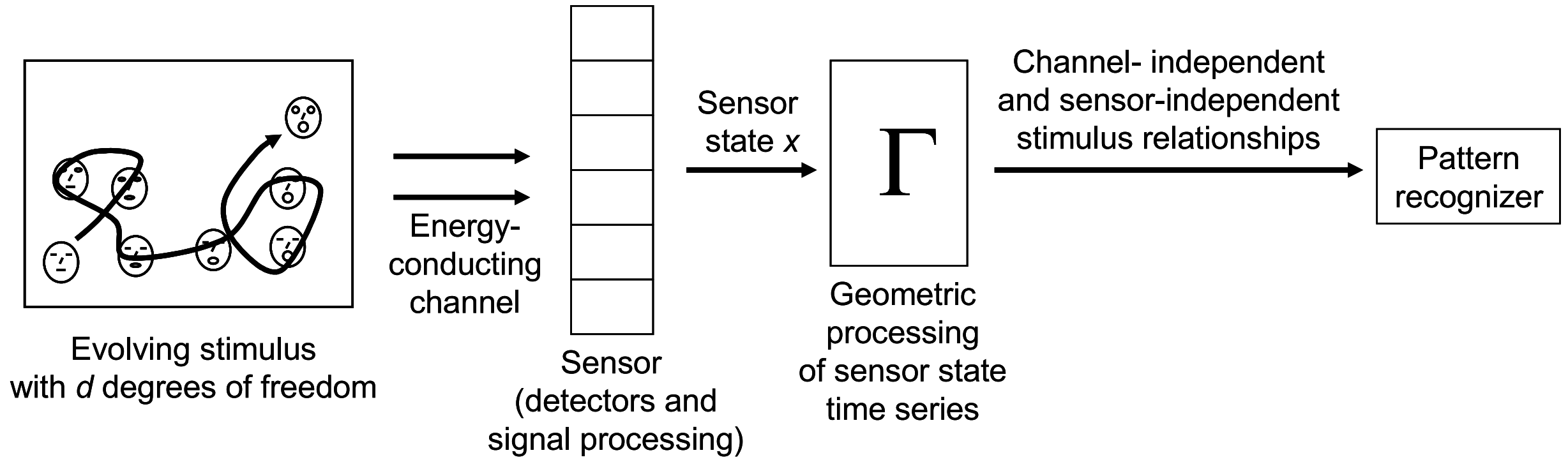

FIG. 3. The architecture of an intelligent sensory device. The energy from the stimulus traverses a channel, before being detected and processed by a sensor. The resulting collection of measurements is used to define a sensor state. The time series of sensor states is processed in order to find channel-independent and sensor-independent relationships among stimulus states, which comprise the input of a pattern recognizer.

Consider a sensory device having the general architecture shown in Figure 3, and suppose that the device is exposed to a "world" consisting of an evolving stimulus. Energy from the stimulus traverses a channel, before being detected and processed by a sensor. This processing defines a sensor state ($x$), which is a function of the stimulus state and is analogous to the state of the brain's primary sensory cortex. In the devices proposed in this paper, the sensor state time series is first processed in order to derive stimulus analogies that are invariant under channel- and sensor-induced transformations of those sensor states. Then, the analogies are used to represent each stimulus state by its channel-independent and sensor-independent location relative to previously-encountered stimulus states, and this information is sent to a pattern recognition module. The same pattern recognition module can be used in two machines equipped with different channels and sensors, because these differences are "filtered out" when the stimulus representation is created.

In a conventional sensory device, the sensor state is sent directly to a pattern recognition module. Because the unprocessed sensor state is channel-dependent and sensor-dependent, the pattern recognition module of each device must be individually "trained". Alternatively, each device's response to stimuli with known configurations must be measured. This calibration is usually done by exposing the device to a special "test pattern" of known stimuli and by using the resulting calibration data (e.g., a transfer function) to explicitly remove the channel- and sensor-dependence of the sensor state before it is sent to the pattern recognizer. In either of these cases, the operator must intervene in order to seize control of the stimulus and in order to take the device "off-line" so that it can be trained or calibrated. Furthermore, the calibration procedure may only be capable of removing linear distortions from the data. In contrast, the sensory devices described in this paper are self-calibrating in the sense that they produce channel-independent and sensor-independent results without any operator intervention, without taking the devices "off-line", and in the presence of non-linear channel- and sensor-related distortions. However, it should be noted that the stimulus representations produced by the new devices only describe the relative locations of stimulus states with respect to one another. Therefore, they may contain less information than the stimulus representations produced by conventional sensory devices, which often describe the absolute configuration of the stimulus in a specific coordinate system.

It is worth noting that previously-reported methods of data representation do not have the channel-independence and sensor-independence of the proposed stimulus representations. For example, in multidimensional scaling[6] and Isomap[7], the representation of stimulus states depends on the assignment of distances between each pair of sensor states, and these distances change if the stimulus measurements are subjected to a transformation, caused by a change of channel or sensors. Likewise, in locally-linear embedding[8], stimulus states are represented by coordinates that are not invariant under all of the stimulus measurement transformations, which are induced by channel and sensor changes. And even less powerful methods of data representation, such as principal components analysis, have the same drawback.

In the next section, we show that the problem of finding channel-independent and sensor-independent relative stimulus locations can be mapped onto the differential geometric problem of finding coordinate-system-independent properties of the trajectory of previously-encountered stimulus configurations. In Section III, we describe large classes of stimulus trajectories for which the desired invariant stimulus representations can be analytically derived. The proposed methods are also illustrated by a numerical simulation in which two machines observe an evolving stimulus through dramatically different channels and sensors. The implications and applications of this approach are discussed in the last section.

## II. THEORY

Suppose that the stimulus is some physical system that has $d$ degrees of freedom, and suppose that the observing machine's sensor makes at least $2d+1$ independent measurements, which are time-independent functions of the stimulus configuration. This means that the channel and sensor are assumed to be stationary, although this assumption can be weakened if the device is run in an adaptive mode (Section IV). The measurement functions define a mapping from the $d$-dimensional stimulus space to a $d$-dimensional subspace of the higher-dimensional ($\geq 2d+1$) measurement space. The Takens embedding theorem, which is well-known in the field of nonlinear dynamics[9], states that this mapping is invertible for almost all choices of measurement functions. Essentially, invertibility is likely because so much "room" is provided by the "extra" dimensions of the higher dimensional space that the $d$-dimensional subspace, which is the range of the mapping, is very unlikely to self-intersect. Suppose that the stimulus evolves so that the stimulus trajectory densely covers a patch of stimulus state space and the corresponding sensor measurements densely cover a patch of the $d$-dimensional measurement subspace. Then, the machine can use dimensional reduction methods[7,8] to learn the location and shape of the measurement subspace, and it can impose an arbitrarily-chosen $x$ ($x_k$, $k = 1, 2, \ldots, d$) coordinate system on it. The quantity $x$ is defined to be the sensor state of the machine, and, according to the embedding theorem, it is invertibly related to the stimulus state. For example, suppose that the stimulus consists of a particle that is moving on a lumpy transparent two-dimensional surface in a laboratory ($d = 2$). Furthermore, suppose that its state is being monitored by three video cameras in various locations and positions, each of which derives two measurements from the imaged scene (e.g., the two pixel coordinates of the particle or two Fourier coefficients of the image or two wavelet coefficients of the image or two …). Notice that the mapping between the particle's position and the measurements of any one camera may not be invertible because of the lumpiness of the surface. As the particle moves over the surface in the laboratory, the six camera measurements move in a two-dimensional subspace within the six-dimensional space of possible measurements, and the location and shape of this subspace can be learned by dimensional reduction of a sufficiently dense set of measurements. Each stimulus configuration (each position of the particle on the lumpy surface) produces a sensor state $x$, which is defined to be the location of the corresponding measurements in some two-dimensional coordinate system on the measurement subspace. The embedding theorem asserts that the mapping between the particle's positions and the sensor states is one-to-one, for almost all choices of camera positions and orientations and for almost all types of image-derived measurements. This is demonstrated in Section III.B, which describes a numerical simulation of exactly this experiment.

In general, our objective is to represent the locations of stimulus states relative to one another in a channel-independent and sensor-independent manner. The discussion in Section I suggests that the machine should learn these relationships from its "life experience", i.e., from the sequence of previously-observed stimuli. In any event, the trajectory of previously-encountered stimuli provides the only possible source of such "structure" on the stimulus space. Because the mapping between the stimulus and sensor states is invertible, the sensor state can be considered to be the location of the stimulus in a particular ($x$) coordinate system on the stimulus space. Therefore, when machines with different channels and sensors observe the same stimulus time series, they are simply using different coordinate systems to view the same trajectory in stimulus state space (Fig. 4). Hence, the problem of defining channel-independent and sensor-independent relative stimulus locations is the same as the problem of finding a coordinate-system-independent description of stimulus locations. Recall that Euclidean geometry provides methods of making statements about a geometric figure that are true in all rotated or translated coordinate systems. Similarly, differential geometry provides the mathematical machinery for making statements about a geometric object (such as the stimulus trajectory) that are true in all linearly and non-linearly transformed coordinate systems[10]. In other words, differential geometry seeks to find properties of a geometric figure that are "intrinsic" to it in the sense that they don't depend on extrinsic factors, such as the coordinate system in which the observer chooses to "view" it (i.e., the choice of numerical labels assigned to the figure's points). This suggests the following strategy: 1) first, use differential geometry to find coordinate-system-independent properties of the trajectory of previously-encountered stimuli; 2) then, use those properties to derive coordinate-system-independent stimulus analogies at each point of the traversed part of the stimulus space; 3) finally, use these analogies as in Fig. 2 to describe relative stimulus locations in a manner that is coordinate-system-independent (and, therefore, channel-independent and sensor-independent).

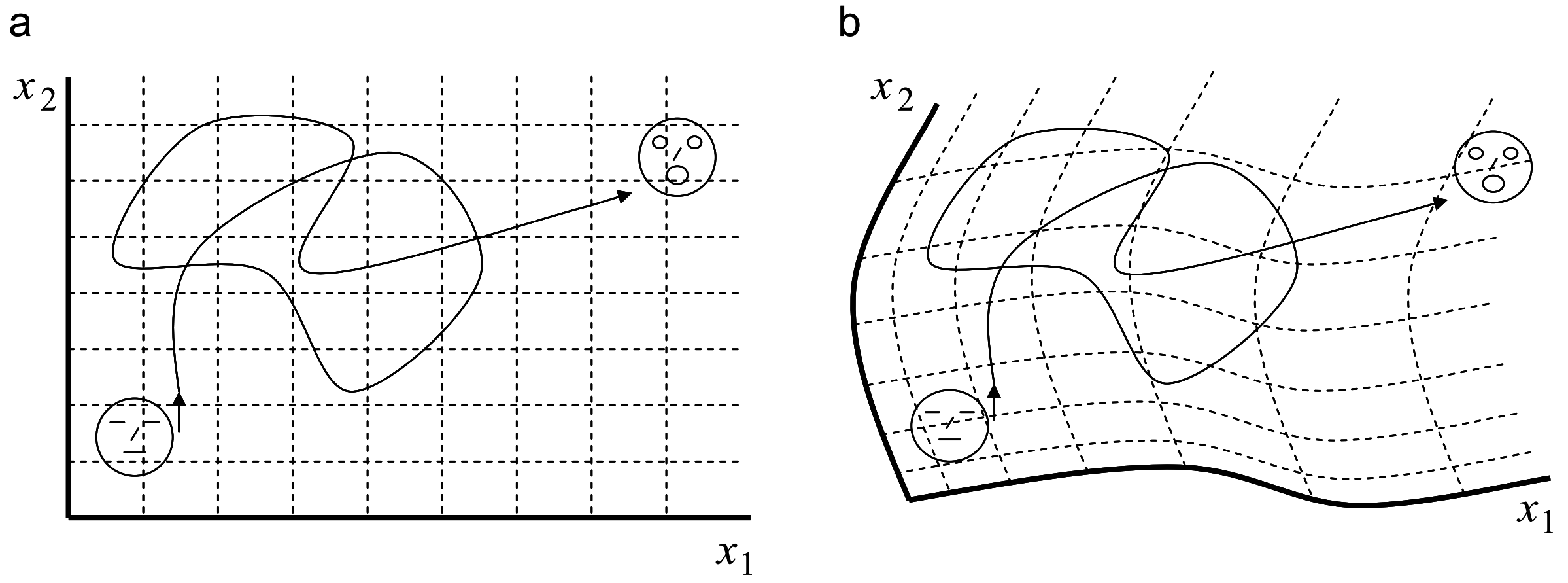

FIG. 4. Because of the embedding theorem, devices with different channels and/or sensors can be considered to be viewing the same stimulus trajectory in different coordinate systems of the stimulus state space (panels *a* and *b*).

From a geometric point of view, a coordinate-system-independent local stimulus analogy specifies a coordinate-system-independent way of moving one short line segment along a second short line segment in order to create a third short line segment. For example, the stimulus analogy in Fig. 1 specifies that, when the line segment $A \rightarrow B$ at $A$ is moved along the line segment $A \rightarrow C$, it becomes the line segment $C \rightarrow D$ at *C*. This is exactly the operation that is performed by the parallel transfer procedure of differential geometry, which comprises a coordinate-system-independent way of moving a vector along a path (Fig. 5). Therefore, if a parallel transfer operation can be derived from the stimulus trajectory, it can be used to define stimulus analogies, and these can be concatenated to define relative stimulus locations in a coordinate-system-independent manner. There are a few methods of using the stimulus trajectory to define a parallel transfer operation[5,11,12], and in this paper, we focus on one of them. First, we use the stimulus trajectory to derive a metric on the stimulus state space (i.e., a coordinate-system-independent method of defining the lengths of line segments and other vectors). Then, we derive a parallel transfer mechanism from that metric by using the following fact: there is a unique parallel transfer procedure that preserves the lengths of transferred vectors with respect to a given metric without producing torsion (without producing reentrant geodesics[10]). In any coordinate system ($x$), this particular parallel transfer operation can be computed in the following manner (Fig. 5): given the metric tensor $g_{kl}(x)$, a vector $V$ at $x$ is parallel transferred along a line segment $\delta x$ into the vector $V + \delta V$ at $x + \delta x$, where

$$\delta V^k = - \sum_{l,m=1,\ldots,d} \Gamma^k_{lm}(x) V^l \delta x_m , \qquad (1)$$

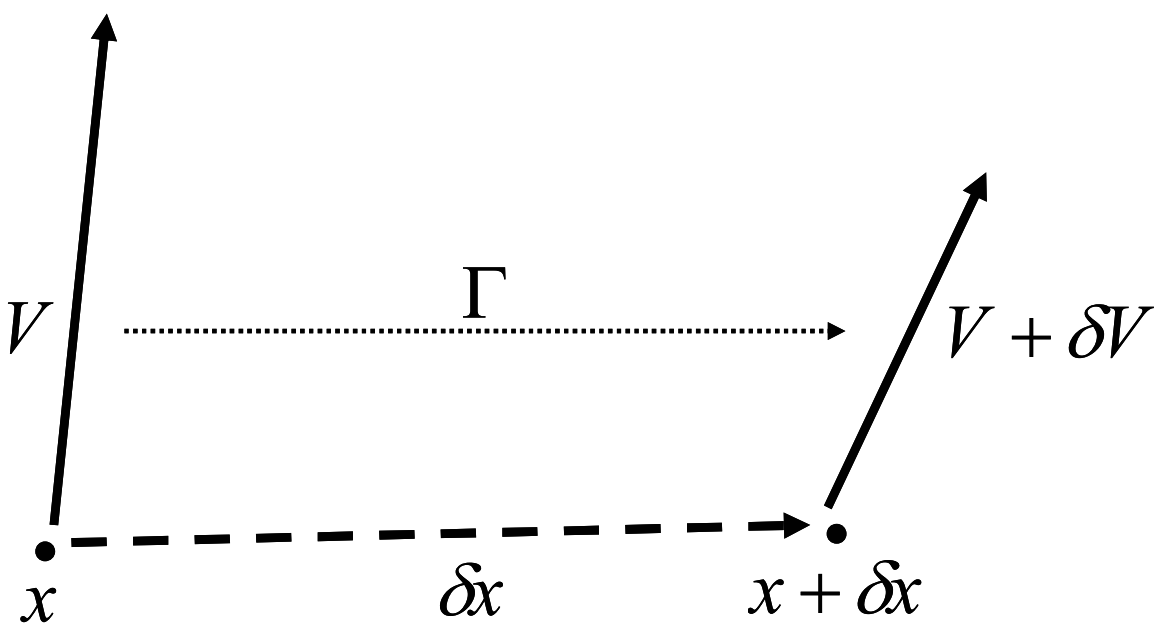

FIG. 5. The parallel transfer procedure $\Gamma$ is a coordinate-system-independent method of moving a vector $V$ at $x$ along the short path segment $\delta x$ to produce an "equivalent" vector $V+\delta V$ at $x+\delta x$.

where $\Gamma^k_{lm}(x)$ (the affine connection at $x$) is given by

$$\Gamma^k_{lm} = \frac{1}{2}\sum_{n=1,\ldots,d} g^{kn}\left(\frac{\partial g_{mn}}{\partial x_l} + \frac{\partial g_{nl}}{\partial x_m} - \frac{\partial g_{lm}}{\partial x_n}\right) \quad (2)$$

and where $g^{kl}$ is the inverse of $g_{kl}$. Therefore, the whole problem of deriving channel-independent and sensor-independent stimulus relationships has been reduced to the task of using the stimulus trajectory to derive a non-singular metric (i.e., a non-singular second rank symmetric covariant tensor). Once that metric has been found, Eqs. 1-2 can be used to define a parallel transfer operation, which makes it possible to define stimulus analogies and relative stimulus locations in a coordinate-system-independent manner (and, therefore, in a channel-independent and sensor-independent manner).

Let $x(t)$ describe the evolving stimulus state in the coordinate system defined by a particular observing machine's channel and sensor. At each point $y$ in stimulus state space, consider the local covariance matrix of the trajectory's "velocity"

$$c^{kl}(y) = \langle \dot{x}_k \dot{x}_l \rangle_{x(t)\sim y} \quad (3)$$

where $\dot{x} = dx/dt$ and the bracket denotes the time average over the trajectory's segments in a small neighborhood of $y$. If this quantity approaches a definite limit as the neighborhood shrinks around $y$, it will certainly transform as a symmetric contravariant tensor. Furthermore, as long as the trajectory has $d$ linearly-independent segments passing near $y$, this tensor can be inverted to form a symmetric covariant tensor, $g_{kl}(y) = (c^{-1})_{kl}$. Thus, Eq. 3 can be used to derive a metric from any stimulus trajectory for which the above-described limit exists. Notice that the metric scales quadratically when time is transformed linearly. However, rescaling time does not change the affine connection (Eq. 2) nor the coordinate-system-independent description of relative stimulus locations.

The right side of Eq. 3 is expected to have a well-defined local limit if the trajectory's local distribution of velocities varies smoothly over the stimulus state space. Specifically, suppose that there is a density function $P(x,\dot{x})$, which varies smoothly with $x$ and which measures the fraction of total trajectory time that the trajectory spends in a small neighborhood $dxd\dot{x}$ of $(x,\dot{x})$ space. In that case, the limit in Eq. 3 exists and is proportional to a second moment of that function.

## III. EXAMPLES

### A. Analytic examples: systems in thermodynamic equilibrium

In this Section, we demonstrate large classes of stimulus trajectories for which the local velocity covariance matrix (and metric) are well-defined and can be computed analytically. Specifically, we construct these trajectories from the behavior of physical systems that can be realized in the laboratory. First, consider a system with $d$ degrees of freedom, and suppose that the system's energy is given by

$$E(x,\dot{x}) = \frac{1}{2}\sum_{k,l=1,\ldots,d} \mu_{kl}(x)\dot{x}_k\dot{x}_l + V(x) \quad (4)$$

where $\mu_{kl}$ and $V$ are some functions and $x$ is the stimulus' location in some coordinate system. Equation 4 is the energy function of a large variety of physical systems in which the degrees of freedom can be spatial and/or internal and in which $d$ may be large or small. The simplest system with this energy is a single particle with unit mass that is moving in potential *V(x)* on a possibly-curved two-dimensional frictionless surface with physical metric $\mu_{kl}$. For example, if the particle is moving on a frictionless spherical, cylindrical, or planar surface in the laboratory, $\mu_{kl}$ is the metric induced on the surface by the Euclidean metric in the laboratory's coordinate system. Returning to the more general case, suppose that the system intermittently exchanges energy with a thermal "bath" at temperature *T*. In other words, suppose that the system evolves along one trajectory from the Boltzmann distribution at that temperature and periodically jumps to another randomly-chosen trajectory from that distribution. After a sufficient number of jumps, the amount of time the particle will have spent in a small neighborhood $dx$ of $x$ and a small neighborhood $d\dot{x}$ of $\dot{x}$ is given by the product of $dxd\dot{x}$ and a density function that is proportional to the Boltzmann distribution[13]

$$\mu(x)\exp\left[-E(x,\dot{x})/kT\right] \tag{5}$$

where $k$ is the Boltzmann constant and $\mu$ is the determinant of $\mu_{kl}$. If this expression is used to evaluate the right side of Eq. 3 and the resulting Gaussian integrals are done, the velocity covariance matrix is found to be well-defined and given by

$$c^{kl}(x) = kT\mu^{kl}(x) \tag{6}$$

where $\mu^{kl}$ is a contravariant tensor equal to the inverse of $\mu_{kl}$. It follows that the trajectory-induced metric on the stimulus space is $g_{kl}(x) = \mu_{kl}(x)/kT$; i.e., it is proportional to the physical metric on the surface. Thus, a metric is induced on the stimulus state space by the trajectories of each member of this large class of physical systems. It follows that Eqs. 1-2 define a coordinate-system-independent method of moving line segments across the stimulus manifold, and this procedure can be used to define channel-independent and sensor-independent stimulus analogies and relative stimulus locations.

### B. Simulated experiment: free particle on a cylindrical surface

As an illustration, consider the following numerical simulation of the scenario described in Section II, in which a particle moved on a transparent frictionless surface and was observed by a machine *Ob* equipped with multiple cameras. Specifically, consider a simulated particle of unit mass that moved on the surface of a cylinder, which was oriented at an arbitrarily-chosen angle in the simulated laboratory coordinate system (Figure 6a). The cylinder was two length units in diameter, and the simulated particle moved on a unit square on the cylinder's surface that was aligned with the cylinder's axis. The particle moved freely (i.e., $V(x) = 0$) in thermal equilibrium with a "bath" for which $kT = 0.01$ in the chosen units of mass, length, and time. The particle's trajectory was created by temporally concatenating 18,274 short trajectory segments that had energies randomly chosen from the corresponding Boltzmann distribution (Eqs. 4-5). The particle was "watched" by *Ob* through three simulated pin hole cameras, which had arbitrarily-chosen positions and faced the cylinder with arbitrarily-chosen orientations (Fig. 6a). The image created by each camera was transformed by an arbitrarily-chosen second-order polynomial, which varied from camera to camera. In other words, each pin hole camera image was distorted by a translational shift, rotation, rescaling, skew, and quadratic deformation that simulated the effect of a distorted optical channel between the cylinder and camera (e.g., thereby simulating a different distorting "goggle" lens in front of each camera). Each camera measured the particle's location in the distorted image on its "focal" plane, as illustrated in Figs. 6b-d. As the particle moved across the cylinder, it created a time series of measurements, each of which consisted of the six measurements made by all three cameras. A dimensional reduction technique (local linear embedding[8]) was applied to this time series in order to identify the underlying measurement subspace, which had two dimensions, and to establish a coordinate system ($x$) on it. The value of $x$ associated with a particular stimulus state (i.e., a particular particle position) defined the sensor state of *Ob*, as described in Section II. Because the stimulus state space (the cylinder surface) had dimensionality $d = 2$ and because the measurement subspace was derived from more than $2d+1$ measurements, the Takens embedding theorem virtually guaranteed that there was a one-to-one mapping between the stimulus and sensor states. Therefore, the $x$ coordinate system on the measurement subspace defined a coordinate system on the stimulus space. The nature of that coordinate system depended on the channels and detectors used to make the machine's measurements (e.g., it depended on the positions, orientations, and optical paths of the three pin hole cameras). Next, Eq. 3 was used to compute the stimulus space's metric in this particular coordinate system. Parallel transfer (Eq. 1) was then used to derive coordinate-system-independent statements about the relationships between various stimuli (i.e., statements about relative stimulus locations as illustrated in Fig. 2). For example, in this experiment, the stimulus position *E* was found to be related to positions *A*, *B*, and $C$ (Fig. 6) by the following procedure: position $E$ was reached by parallel transferring the increment $A \rightarrow C$ along itself 4.8 times, followed by the parallel transfer of the local equivalent of the increment $A \rightarrow B$ along itself 29.9 times.

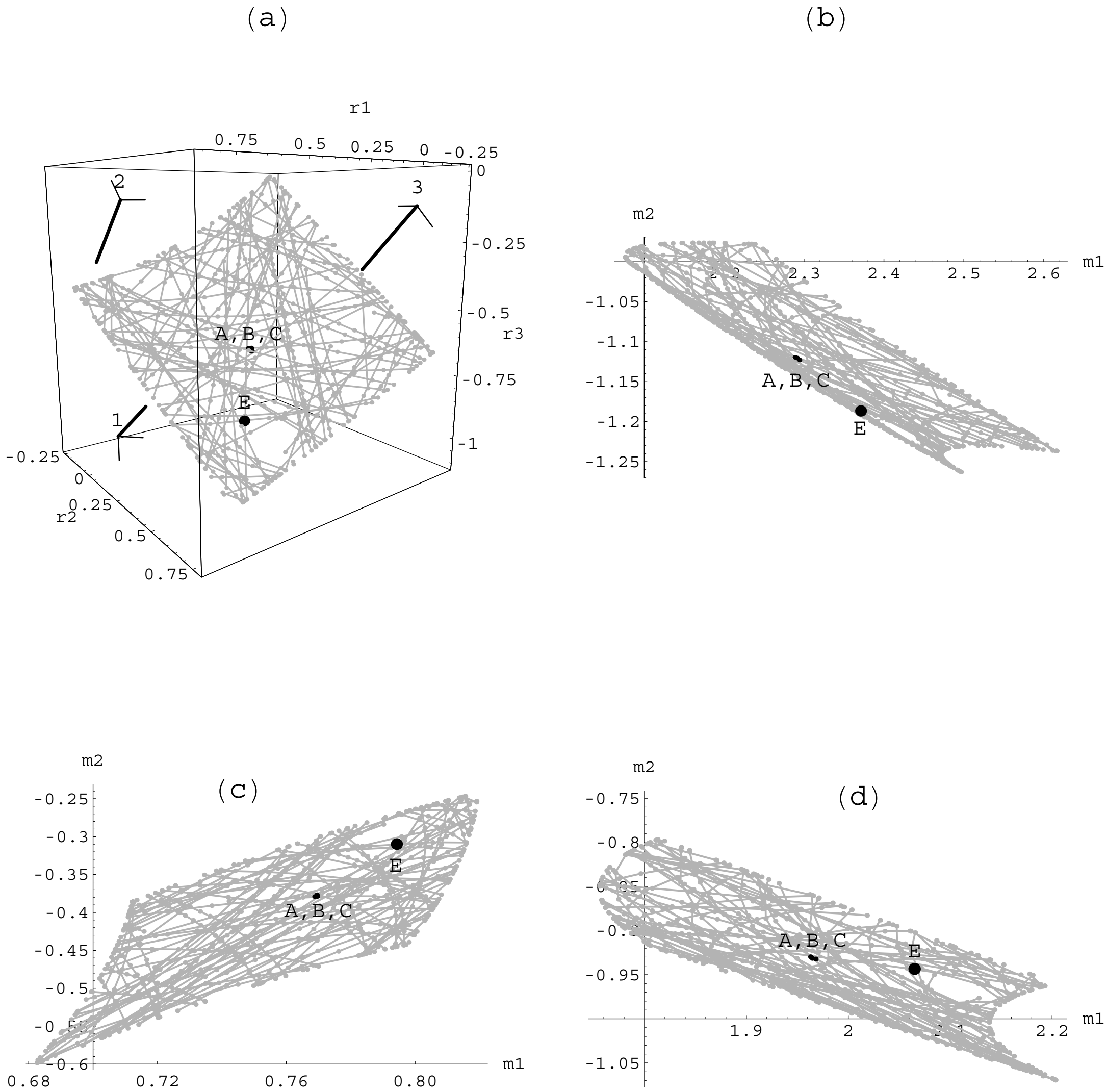

FIG. 6. Simulated experiment in which machine *Ob* used three "goggled" pin hole cameras to observe a particle moving on a transparent cylindrical surface. (a) Some of the 18,274 observed particle trajectory segments are shown in the "laboratory" coordinate system. The three sets of orthogonal axes show the orientation of each camera's image plane (two short thin lines) and the normal to it (long thick line). (b) – (d) Plots of each camera's output signal, which consisted of the location of the particle on its image plane, after a quadratic transformation was performed in order to simulate the effects of distorting "goggles". The dots along the trajectory segments are separated by equal time intervals. *Ob* used the resulting time series of six measurements at each time point in order to compute the location of test point *E* relative to points *A, B,* and *C.*

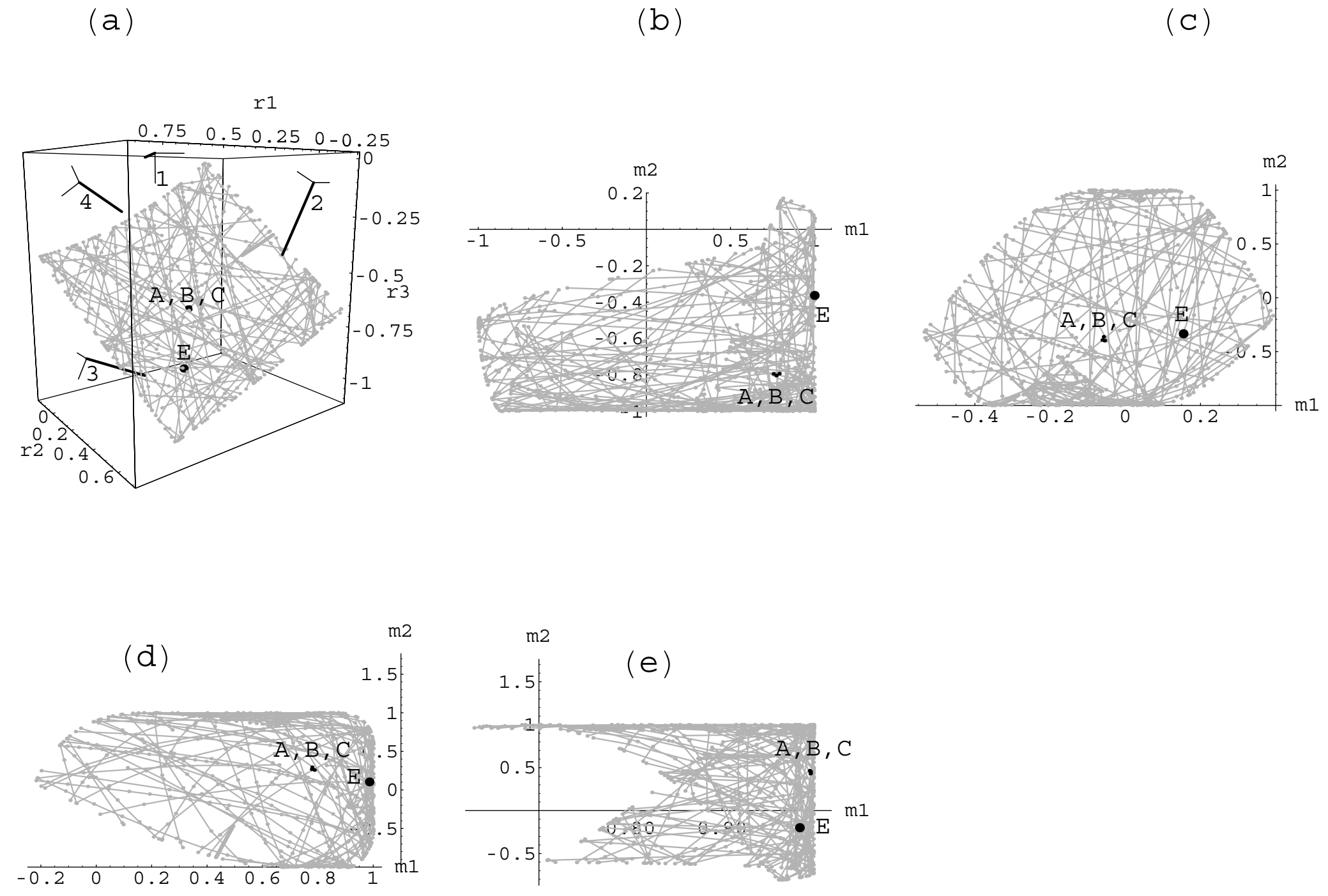

FIG. 7. Simulated experiment in which machine $Ob'$ used four "Fourier" pin hole "cameras" to observe a particle moving on a transparent cylindrical surface. (a) Some of the 17,674 observed particle trajectory segments are shown in the "laboratory" coordinate system. The four sets of orthogonal axes show the orientation of each camera's image plane (two short thin lines) and the normal to it (long thick line). (b) – (e) Plots of each camera's output signal, which consisted of the real part of one Fourier coefficient and the imaginary part of another Fourier coefficient of the image of the particle. The dots along the trajectory segments are separated by equal time intervals. $Ob'$ used the resulting time series of eight measurements at each time point in order to compute the location of test point $E$ relative to points $A$, $B$, and $C$.

A second machine $Ob'$ was equipped with *four* pin hole "cameras" with arbitrarily-chosen positions and orientations that differed from those of the three cameras of machine $Ob$ (Fig. 7a). However, unlike $Ob$, each "camera" of $Ob'$ did *not* record the focal plane location of the imaged particle. Instead, each of $Ob'$'s cameras recorded two measurements, $\cos(\vec{k}_1 \cdot \vec{y})$ and $\sin(\vec{k}_2 \cdot \vec{y})$, where $\vec{y}$ was the focal plane location of the imaged particle and where $\vec{k}_1$ and $\vec{k}_2$ were two wave vectors that were arbitrarily chosen and were different for each camera. In other words, the output of each of $Ob'$'s camera (Figs. 7b-e) was the real part of one Fourier coefficient of the particle's image, together with the imaginary part of another Fourier coefficient. Notice that the measurements of any one camera were not in one-to-one correspondence with the particle's position on the cylinder (Figs. 7b-e). This is because some cameras viewed portions of the cylindrical surface through other parts of the transparent cylinder and because the "Fourier" measurement functions oscillated across each camera's field of view. $Ob'$ "watched" the particle as it traversed 17,674 short trajectory segments that were randomly-chosen from the above-described Boltzmann distribution (Fig. 7a). As the particle moved across the cylinder, $Ob'$ recorded a time series of measurements, each of which consisted of the *eight* measurements made by its four "Fourier" cameras. As before, dimensional reduction was applied to this time series in order to identify the underlying 2D measurement subspace and to establish a coordinate system ($x$) on it. Even though individual camera measurements were not in one-to-one correspondence with the particle's position, the embedding theorem guaranteed that the measurement subspace was invertibly related to the stimulus configuration. Therefore, the $x$ coordinate system defined a coordinate system on the stimulus space. As before, the nature of that coordinate system depended on the channels and detectors used to make the machine's measurements (e.g., it depended on the positions and orientations of the four cameras, as well as on the $(\cos(\vec{k}_1 \cdot \vec{y}), \sin(\vec{k}_2 \cdot \vec{y}))$ transformations used to define their output). Next, the stimulus space's metric was computed in this particular coordinate system, and parallel transfer was used to derive coordinate-system-independent statements about the relationships between various stimuli. For example, machine *Ob*' found that the four previously-considered particle positions (*A*, *B*, *C*, and *E*) were related by the following procedure: position *E* was reached by parallel transferring the increment $A \rightarrow C$ along itself 8.2 times, followed by the parallel transfer of the local equivalent of the increment $A \rightarrow B$ along itself 29.0 times. Thus, *Ob* and *Ob*' were in near agreement about the relative locations of these four stimuli, even though they used radically different channels and sensors to observe the stimuli and even though they were not calibrated and did not communicate in any way.

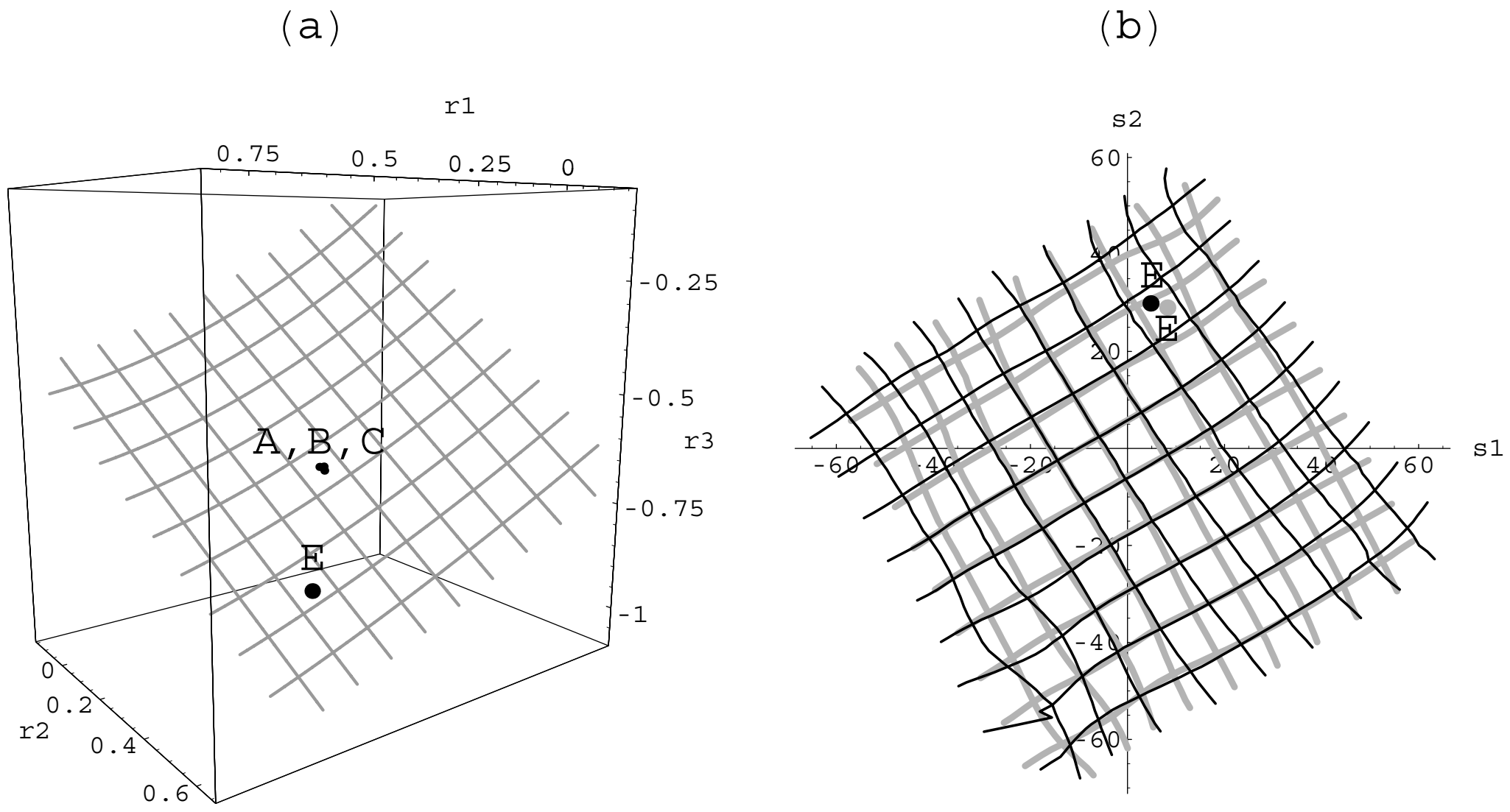

FIG. 8. "Maps" of the stimulus space created by *Ob* and *Ob*' from the simulated data in Figs. 6b-d and Figs. 7b-e, respectively. (a) The grid-like array of test points (plus an additional test point *E*), which each device located with respect to points *A*, *B*, and *C*, is shown in the "laboratory" coordinate system. (b) The thin black lines and thick gray lines show the relative locations of the grid-like array of test points with respect to *A*, *B*, and *C*, computed by *Ob* and *Ob*', respectively. The black and gray dots show the corresponding relative locations of test point *E*. The points *A*, *B*, and *C* are located at *(0,0)*, *(0,1)*, and *(1,0)*, respectively. A given test point was plotted at $(s_1, s_2)$ if that point was reached by starting at *A*, parallel transferring the segment *AC* along itself $s_1$ times, and then parallel transferring the local equivalent of the segment *AB* along itself $s_2$ times.

Figure 8 shows that the two machines were also in near agreement about the locations of many other particle positions with respect to *A*, *B*, and *C*. Figure 8a depicts a set of "test" points (like *E*) that formed a grid on the cylindrical stimulus space. Each point on this grid was used to induce a sensor state in *Ob*, which used its parallel transfer procedure to compute the location of that sensor state with respect to the sensor states produced by points *A*, *B*, and *C*. Suppose that the test point was reached by parallel transferring the increment $x_A \rightarrow x_C$ along itself $s_1$ times, followed by the parallel transfer of the local equivalent of $x_A \rightarrow x_B$ along itself $s_2$ times. Then, a small black point was plotted at coordinates $(s_1, s_2)$ in Fig. 8b. Thus, the resulting thin black lines in Fig. 8b shows where *Ob* "perceived" the grid points to be located with respect to *A*, *B*, and *C*. Similarly, the thick gray lines in Fig. 8b show where *Ob*' "perceived" the grid-like array of test points to be located. The near coincidence of the thin black lines and thick gray lines demonstrates that *Ob* and *Ob*' were in near agreement about the location of each grid point with respect to the locations *A*, *B*, and *C*. It should be emphasized that the grid in Fig. 8 was *not* used to calibrate the two devices; no such calibration procedure was performed. Rather, the grid is just a convenient array of individual test points that the two devices were "asked" to locate in a completely independent manner.

In this example, the "world" was a simple physical system with two degrees of freedom: namely, a free particle moving on a transparent frictionless cylindrical surface in equilibrium with an invisible thermal bath. *Ob* and *Ob*' observed sample stimuli in that world using "eyes" that were radically different (perhaps, analogous to the radical difference between human and insect eyes). The signal processing units of the two devices could be considered to be "locked inside their chassis", much like human brains are "locked inside their skulls". These processing units had no knowledge whatsoever of the world "out there", except for the measurements streaming from their sensors (six and eight measurements per unit time for *Ob* and *Ob*', respectively). For example, they certainly did not know that they were observing a particle moving on a cylinder, rather than some other physical system. Furthermore, *Ob* and *Ob*' had absolutely no information about the channels and sensors that were producing the stream of measurements. For instance, they had no knowledge of the number of cameras, no information about the camera positions and orientations, no knowledge of the "goggle" lenses on some of the cameras, no information about what image features were measured (e.g., pixel positions vs. Fourier coefficients vs. …). In fact, they did not even know what type of energy was being detected (e.g., optical vs. radio frequency vs. acoustic vs. …). Nevertheless, both devices independently deduced that the observed stimulus had two degrees of freedom, and they independently derived a parallel transfer procedure on the two-dimensional stimulus space. Then, when the observing devices were queried about the relative locations of four stimuli (e.g., *A, B, C*, and *E*, as in Fig. 2), they used parallel transfer to produce nearly the same result. This concordance can be traced to the fact that both devices observed trajectory segments randomly drawn from the same statistical distribution. This guaranteed that they derived velocity covariance matrices, metrics, and parallel transfer procedures that were the same, although they were expressed in different machine-specific coordinate systems. These could then be used to represent the locations of stimuli with respect to one another in a coordinate-system-independent way (and, therefore, in a machine-independent way). All of this was done in a completely "blind" fashion: i.e., without explicitly calibrating either machine's response to stimuli and without any prior knowledge of the physical nature of the stimulus. Of course, for this reason, these machines were not able to recover the absolute laboratory coordinates of the particle. In fact, they were not even able to learn that the stimulus consisted of a moving particle, rather than some other physical system.

The *s* coordinate system in Fig. 8 has the following differential geometric interpretation. It is known that the affine connection at any given point vanishes in certain special "geodesic" coordinate systems, which differ by arbitrary affine transformations. Furthermore, at any manifold point, the metric is proportional to the identity matrix in a special "canonical" subset of these locally geodesic coordinate systems (the members of that subset differing from one another by translations, rotations, and uniform scaling). For any choice of points *A*, *B*, and *C*, the coordinate system defined by the procedure in Fig. 2 (i.e., the *s* coordinate system) is locally geodesic at *A*. Because the *AB* and *AC* line segments in Fig. 8a were chosen to be orthogonal and to have equal length, the *s* coordinate system in Fig. 8 is one of these canonical geodesic coordinate systems at *A*.

### C. Analytical example: moving video camera

As another example, consider a car that is being driven through a city, which is laid out on a flat surface, and suppose that the car is equipped with a video camera that is always pointing west. Then, the stimulus space is two-dimensional ($d = 2$), being comprised of the collection of western views associated with all locations in the city. Suppose that the camera's output consists of five properties of each image encountered by the car along its route. The time series of these measurements sweeps out a trajectory within a two-dimensional subspace of the five-dimensional space of possible measurements. If the car's path through the city is sufficiently dense, the trajectory of measurements will densely cover this two-dimensional subspace, and the machine can use dimensional reduction methods in order to learn the subspace's location and shape. The machine's sensor state is defined to be the measurement's location in some two-dimensional coordinate system on this subspace. Because the stimulus and sensor states are invertibly related according to the embedding theorem, the sensor state can be taken to define the location of the stimulus in the machine-defined coordinate system on the stimulus space. Now, suppose that the car is being driven at a constant speed $v$ along randomly-chosen directions in the Euclidean coordinate system of the city. Then, in this coordinate system, the time average in Eq. 3 is well-defined and equal to $c^{kl}(x) = v^2 \delta_{kl} / 2$ where $\delta_{kl}$ is the Kronecker delta, and the trajectory-derived metric on the stimulus space is proportional to the Euclidean metric, namely $g_{kl}(x) = 2\delta_{kl} / v^2$. In any other coordinate system (e.g., in the coordinate system defined by the machine's sensor state), the trajectory-derived metric will be similarly proportional to the form of the Euclidean metric in that coordinate system. This implies that two different machines will derive the same metric (each in its machine-defined coordinate system). Therefore, they will agree on all statements about relative stimulus locations (i.e., all statements about the geography of the city), despite the fact that they may be equipped with different cameras (e.g., cameras with different lenses or "goggles"), they may use different image-derived measurements, etc. It is interesting to note that, if the car drives more slowly in one part of the city than in others, the trajectory-derived metric will be larger there, and the machine will "perceive" that part of the city to be relatively larger.

In this Section, we demonstrated large classes of trajectories that endowed the stimulus manifold with a metric (and, consequently, with channel-independent and sensor-independent stimulus relationships). It is likely that metrics can be derived from many other trajectories, especially those with local velocity distributions that vary sufficiently smoothly across stimulus space. However, it is an experimental question whether this is the case for the trajectories of other specific physical stimuli. For example, experimental results suggest that a metric can be derived from the stimulus trajectories produced by spoken English. In these experiments, the metric was used to derive a channel-independent representation of speech trajectories, with the ultimate goal of achieving channel-independent automatic speech recognition. A complete description of these results is available in reference 12.

## IV. DISCUSSION

We have shown how a machine, which observes stimuli through an uncharacterized channel and sensor, can glean machine-independent information (i.e., channel- and sensor-independent information) about the relative locations of stimulus configurations. This is possible if the following two conditions are satisfied by the observed "world" and by the observing device, respectively: 1) the "world's" stimulus trajectory has a well-defined local velocity covariance matrix; 2) the observing device's sensor state is invertibly related to the stimulus state. The first condition guarantees that the stimulus trajectory endows the stimulus space with a metric and parallel transfer procedure, which can then be used to represent relative stimulus locations in a coordinate-system-independent manner. As shown in Section II, this requirement is satisfied by a large variety of physical systems, and, in general, it is expected to be satisfied by stimuli with velocity distributions varying smoothly across stimulus state space. The second condition means that the machine defines a specific coordinate system on the stimulus state space, with the nature of that coordinate system depending on the machine's channels and detectors. Thus, machines with different channels and sensors "see" the same stimulus trajectory, but in different machine-specific coordinate systems. Because of the Takens embedding theorem, this requirement is almost certainly satisfied by any device that measures more than *2d* independent properties of the stimulus, where $d$ is the number of stimulus degrees of freedom. Taken together, the two conditions guarantee that the observing device can record the "world's" trajectory in its machine-specific coordinate system and then derive coordinate-system-independent (and, therefore, machine-independent) representations of the relative locations of stimulus configurations. The resulting description of a stimulus is an "inner" property of its trajectory in the sense that it does not depend on extrinsic factors like the observer's choice of a coordinate system in which the stimulus is viewed (i.e., the observer's choice of channels and sensors). In other words, the resulting description is an intrinsic property of the temporal evolution of the "real" stimulus that is "out there" broadcasting energy to the observer. In contrast, the stimulus properties measured by a machine's detectors (and the corresponding dimensionally-reduced sensor state $x$ of the device) comprise the "private experience" of that particular machine, which may not be shared by a different machine.

Like a human, such a machine represents the world in a way that depends on statistical properties of its past sensory "experience". Two such machines, using different channels and sensors, will represent stimulus relationships in the same way, as long as their previous "experiences" imposed the same geometry on stimulus space, and this will be true as long as the velocity covariance matrices of their previously-observed trajectories are identical when expressed in the same stimulus coordinate system. This was illustrated by the numerical experiment in Section III.B, in which the two observing machines had *different* "life experiences" consisting of different sets of random samples from a single statistical distribution of stimulus trajectories. Nevertheless, because the samples were drawn from the same statistical distribution, the two observers derived the same velocity covariance matrices, the same stimulus space geometry, and the same stimulus representations. On the other hand, if two devices have experienced stimulus trajectories with different velocity covariance matrices (e.g., trajectories drawn from different statistical distributions), they will not agree on all statements about relative stimulus locations. For example, there will be striking disagreements if the trajectory-derived metric of one machine is flat and the trajectory-derived metric of the other one is curved[1].

As discussed in Section I, the methodology of this paper should not be confused with conventional ways of calibrating instruments. The usual goal of such calibration is to enable a machine to measure the stimulus configuration in a specific coordinate system. This requires an operator take the device "off-line" in order to measure its response to a series of test stimuli with known configurations in that coordinate system (e.g., measure the device's transfer function). The technique described in this paper does not attempt to recover the absolute stimulus configuration in any particular coordinate system. Rather, it is designed to determine properties of stimuli (i.e., the relative locations of their configurations) that are valid in *any* coordinate system in stimulus space. This can be achieved without using conventional test patterns and calibration tables to normalize the device's sensor states. Instead, they can be normalized with respect to the metric induced on stimulus space by a statistical property of the evolution of the observed "world" itself (i.e., the local velocity covariance matrix of the stimulus time series). The preceding paragraph described a disadvantage of this approach: if two machines have been exposed to statistically-different samples of the "world's" stimulus time series, they may describe the same stimuli in different ways.

The methodology in this paper should also be compared to the way physicists describe physical systems in a coordinate-system-independent manner. A particle's equation of motion is typically written in a coordinate-system-independent manner by introducing the metric of the manifold on which it moves. The metric is determined by fitting the equations of motion to the observed motion of a test particle or, in general relativity, by finding the metric that is produced by the distribution of matter and energy in the manifold. The resulting metric can then be used to derive a coordinate-system-independent description of the particle's trajectory, as described above. In this paper, the metric is derived from a statistical property of the trajectory of the stimulus. This method is useful in a different set of circumstances than is the conventional physical method. For example, one need not know the form of the equations of motion of the stimulus; in fact, the stimulus need not even obey an equation of motion.

Up to now, we have assumed that the channel and sensor of a machine are time-independent and that a machine derives a parallel transfer mechanism from the history of all previously-encountered stimuli. However, if a machine derived its affine connection from stimuli encountered in the most recent time interval $\Delta T$, it could adapt to channels and/or sensors that drift over a longer time scale. It would continue to represent relative stimulus locations in the same way, as long as the local velocity covariance matrices of the stimulus trajectory were temporally stable. Notice that the trajectories constructed in Section III have this kind of temporal stability. This process of adaptation, which is analogous to human adaptation in the goggle experiments, was demonstrated with simple experimental examples in reference 14.

It is interesting to consider the kinds of "memories" that one of these devices could have. A neural network or parametric method could be used to store the location and shape of the *d*-dimensional measurement subspace, once the machine has learned that information by observing a sufficiently dense stimulus trajectory. This subspace contains all sensor state measurements encountered in the "world". The machine could also store the affine connection, which encodes the relative locations of sensor states in that subspace. For example, the weights of a neural network could be adjusted so that it comprises a kind of "connection engine" for mapping each sensor state $x$ onto $\Gamma^k_{lm}(x)$. If the machine subsequently re-experiences a sensor state (e.g., the sensor state corresponding to stimulus *E* in Fig. 2), it can use the stored affine connection to deduce it's location relative to other nearby sensor states in the measurement subspace (e.g., *A*, *B*, and *C* in Fig. 2). Thus, previously-experienced stimulus relationships can be recomputed at any time, as long as the machine's memory contains the previously-experienced affine connection. On the other hand, suppose that the parallel transfer mechanism has changed since a stimulus was first encountered, because the machine has experienced a statistically different stimulus trajectory in the intervening time. Then, the machine will describe the relative location of a stimulus differently than it did before. Likewise, the trajectory segment of a recalled train of events may be described differently with the new affine connection than it was with the affine connection that existed at the time those events first transpired. Despite these differences, each description of the events is valid at the time it was derived, in the sense that it is coordinate-system-independent (and, therefore, independent of factors extrinsic to the stimulus trajectory, such as the nature of the observing machine's channel and sensors).

The nonlinear signal processing method presented in this paper could be used as a representation "engine" in the "front end" of an intelligent sensory device[15]. It would produce channel-independent and sensor-independent stimulus locations that could be used by the device's pattern recognition module in order to recognize stimuli. Because the effects of channels and sensors have been "filtered out" of these representations, devices with different channels and sensors could use the same pattern recognition module, without recalibrating their detectors or retraining the recognizer. As suggested by the example in Section III.C, this may make it possible to design detector-independent computer vision devices. Furthermore, because the method does not explicitly depend on the nature of the detectors, it is a natural way to achieve multimodal fusion of audio and optical sensors. Finally, there is the possibility of creating a speech-like telecommunications system that is resistant to channel-induced corruption of the transmitted information[12]. In such a system, information is carried by the coordinate-system-independent relationships among the transmitted power spectra.

It is philosophically interesting that the channel-independent and sensor-independent information gleaned by these machines reflects the intrinsic properties of the observed *trajectory* of the stimulus state, not the absolute properties of the physical stimulus itself. As an illustration, consider a machine that has been exposed exclusively to a "world" consisting of certain physical stimuli (e.g., a moving particle on a two-dimensional surface) and has used the trajectory of those stimuli to learn their coordinate-system-independent relative locations. Now, consider a second machine, remotely-located in another room or another galaxy, that has been exposed exclusively to a different "world" of physical stimuli (e.g., a time series of changing faces, like those in Fig. 4). If the two stimulus spaces are related by a one-to-one mapping that maps the first stimulus trajectory onto the second one, the first machine will describe the relative locations of its stimuli exactly as the second machine describes the relative locations of the corresponding states of its stimulus. Suppose that each machine has not been exposed to any other stimuli and that it only has access to information about the relative locations of observed stimuli (i.e., it is "unaware" of the raw sensor measurements for each stimulus). Then, the two machines will not be able to discern any differences between their "worlds", even if they freely communicate observed stimulus relationships to one another. These devices will only discover the differences between their stimuli if each one measures the locations of its stimuli relative to common "reference" stimuli that both devices are able to observe. In other words, because the stimulus representation is simply a property of the stimulus trajectory, many different physical systems will lead to the same stimulus representation if their trajectories are isomorphic. Essentially, this method maps "reality" onto representations in a many-to-one way.

The remarkable channel-independence and sensor-independence of human perception have been the subject of philosophical discussion since the time of Plato (e.g., the allegory of "The Cave"), and these issues have also intrigued modern neuroscientists. Given that we are "locked inside our own heads", how can we construct a valid representation of the "real" stimuli that are "out there"? Why do different people tend to have similar perceptions despite significant differences in their sensory organs and neural processing pathways? How can an individual perceive the intrinsic constancy of a stimulus even though its appearance is varying because of changing observational conditions? This paper shows how to design sensory devices that invariantly represent stimuli in the presence of processes that transform their sensor states. These stimulus representations are invariant because they encode "inner" properties of the time series of stimulus configurations themselves; i.e., properties that are independent of the observer's choice of channels and sensors. Significant evolutionary advantages would accrue to organisms that developed the ability to represent sensory information in this "machine-independent" way. For instance, they would not be confused by the appearance-changing effects of different observational conditions, and communication among different organisms would be facilitated because they independently represent stimuli in the same way. Biological experimentation is required to determine whether humans and other species achieve these objectives by means of the general approach in this paper. However, it would not be surprising if they did, because there are not many other ways of representing the world in an invariant fashion.

## ACKNOWLEDGMENTS

The author is indebted to Michael A. Levin for numerous informative discussions and insightful comments over the years. The partial support of the National Institute of Deafness and Communicative Disorders is also gratefully acknowledged.